\documentclass[runningheads]{llncs}
\usepackage{graphicx}

\usepackage{tikz}
\usepackage{comment}
\usepackage{amsmath,amssymb} 
\usepackage{color}

\usepackage[accsupp]{axessibility}  

\usepackage{subfigure}
\usepackage{cite}
\usepackage[colorlinks=true]{hyperref}
\usepackage{algpseudocode}
\usepackage[ruled, linesnumbered]{algorithm2e}
\usepackage{booktabs}
\usepackage{multirow}
\usepackage{bbm}
\usepackage{pifont}
\newcommand{\cmark}{\ding{51}}%

\newcommand\blfootnote[1]{%
\begingroup
\renewcommand\thefootnote{}\footnote{#1}%
\addtocounter{footnote}{-1}%
\endgroup
}

\begin{document}
\pagestyle{headings}
\mainmatter
\def\ECCVSubNumber{4992}  

\title{Prototypical Contrast Adaptation for Domain Adaptive Semantic Segmentation} 

\titlerunning{Prototypical Contrast Adaptation for Domain Adaptive Segmentation}
%
\author{Zhengkai Jiang\inst{1} \and 
Yuxi Li\inst{1} \and Ceyuan Yang\inst{2} \and
Peng Gao\inst{2} \and \\ Yabiao Wang\inst{1\dagger} \and
Ying Tai \inst{1} \and
Chengjie Wang\inst{1\dagger}}
\authorrunning{Zhengkai Jiang et al.}
%
\institute{Tencent Youtu Lab \and
The Chinese University of Hong Kong \\
\email{\{zhengkjiang, caseywang\}@tencent.com}}
\maketitle
\begin{abstract}
Unsupervised Domain Adaptation (UDA) aims to adapt the model trained on the labeled source domain to an unlabeled target domain. In this paper, we present Prototypical Contrast Adaptation (ProCA), a simple and efficient contrastive learning method for unsupervised domain adaptive semantic segmentation. 
Previous domain adaptation methods merely consider the alignment of the intra-class representational distributions across various domains, while the inter-class structural relationship is insufficiently explored, resulting in the aligned representations on the target domain might not be as easily discriminated as done on the source domain anymore. 
Instead, ProCA incorporates inter-class information into class-wise prototypes, and adopts the class-centered distribution alignment for adaptation. By considering the same class prototypes as positives and other class prototypes as negatives to achieve class-centered distribution alignment, ProCA achieves state-of-the-art performance on classical domain adaptation tasks, {\em i.e., GTA5 $\to$ Cityscapes \text{and} SYNTHIA $\to$ Cityscapes}. Code is available at \href{https://github.com/jiangzhengkai/ProCA}{ProCA}.
\blfootnote{ $^{\dagger}$Corresponding author.}
\end{abstract}

\section{Introduction}

Semantic segmentation is a fundamental computer vision task, which requires per-pixel predictions for a given image. Recently, with the development of deep neural networks (DNN)~\cite{he2016deep,huang2017densely,wang2020deep,jiang2019video,jiang2020learning,jiang2022stc,xu2022dirl}, semantic segmentation has achieved remarkable progress~\cite{long2015fully,chen2017deeplab,zhao2017pyramid}. However, state-of-the-art methods still suffer from significant performance drops when the distribution of testing data is different from training data owing to the domain shifts problem ~\cite{luo2019taking, pan2020unsupervised, mei2020instance}. At the same time, labeling pixel-wise large-scale semantic segmentation in the target domain is time-consuming and prohibitively expensive. Thus, Unsupervised Domain Adaptation (UDA) is a promising direction to solve such problem by adapting a model trained from 
largely labeled source domain to an unlabeled target domain without additional cost of annotations. 

\begin{figure}
\centering
\includegraphics[width=0.7\textwidth]{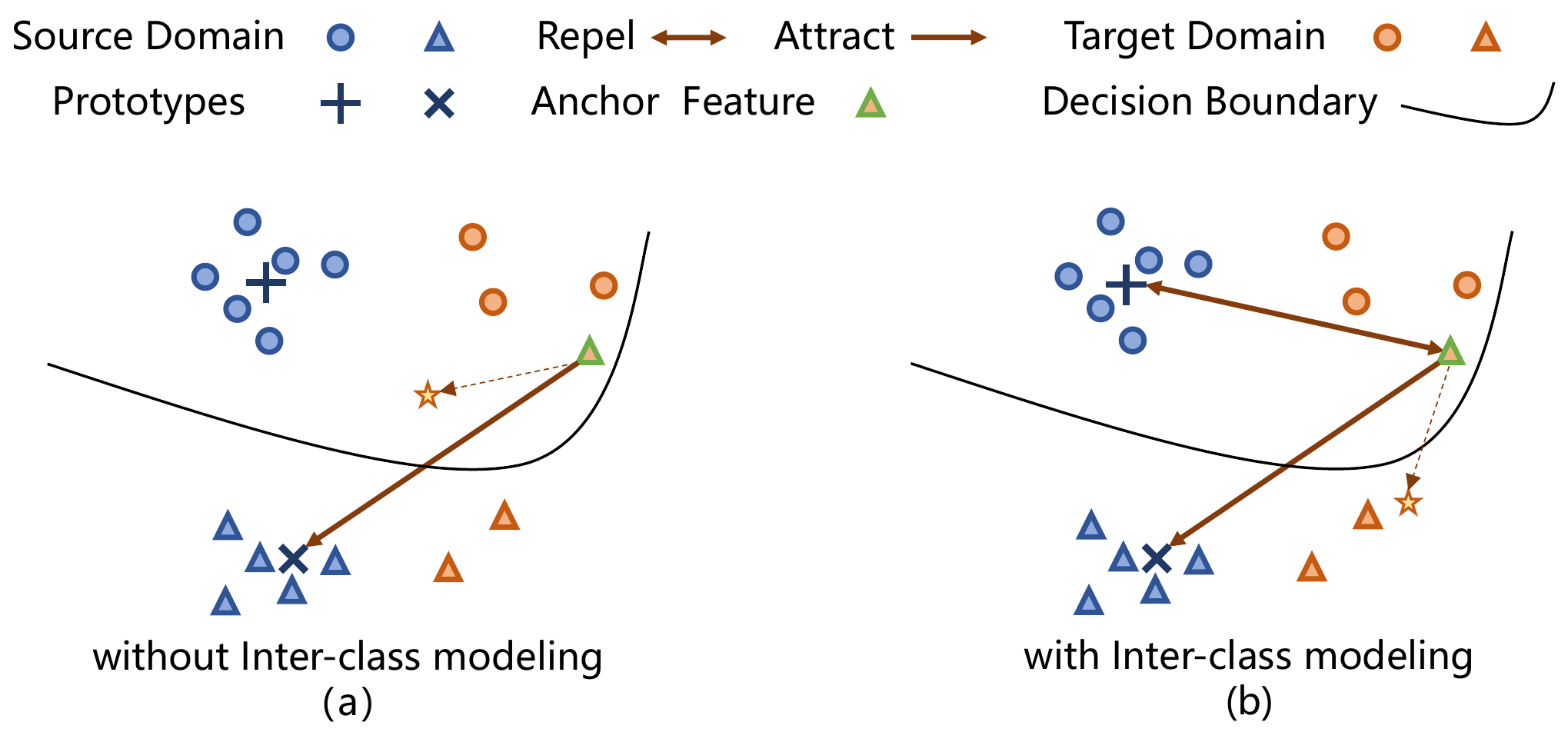}
\caption{Illustration of inter-class modeling. $\star$ means the adapted feature of target domain. With explicit inter-class constraints during adaptation, adapted features of target domain can appear at the right place of decision boundary.}
\label{matching}
\end{figure}

Several works relying on adversarial training~\cite{tsai2018learning, vu2019advent, hoffman2018cycada} have achieved remarkable progress for UDA semantic segmentation. These methods reduce the domain discrepancy between source and target domains by minimizing a series of adversarial training losses. Specifically, it is formulated as a two-player game, where a backbone network ({\em i.e. ResNet-101 backbone}) serves as the feature extractor, while a discriminator identifies which domain the features are derived from. To reach equilibrium in this minmax game, it requires the backbone network to produce the domain invariant representations for generalization. 
Such adversarial training will result in 
aligned and indistinguishable feature distributions between two domains. However, even though the global feature distributions across domains become closer, it is not guaranteed that pixels attributing to different semantic categories in the target domain are well separated, leading to poor generalization ability and even inferior performance.

To tackle the issues above, some works attempt to take the category-wise information into account. 
The idea of encouraging high-confidence predictions is achieved by minimizing the entropy of the output~\cite{vu2019advent}. The discrepancies between the outputs of two classifiers are utilized to achieve category-level alignment implicitly~\cite{luo2019taking}. In addition, a fine-grained adversarial learning framework~\cite{wang2020classes} is proposed to incorporate class information into domain discrimination, which helps to align features at a fine-grained level. However, prior approaches tend to apply such adversarial training in the intra-class, without considering the consistency of the representational structure between the source and target domains. Namely, to some extent, multiple categories on the target domain could be projected to a same group, which are usually well-discriminated on the source domain on the contrary. Therefore, merely considering the intra-class distributional alignment might be insufficient to 
make the best of the learned representations from labeled source data.

In order to fully exploit the class-level information, we propose \emph{Prototypical Contrast Adaptation} (ProCA) for unsupervised domain adaptive semantic segmentation. 
Intuitively, the same category on different domains is supposed to share the high representational similarity. Therefore, multiple prototypes, \emph{i.e.}, the approximated representational centroid of various categories are utilized to depict the inter-class relationship for both source and target domains. 
Specifically, after acquiring the segmentation model trained only on the source domain, category-wise prototypes features are obtained by calculating the centroids of features on the source domain. Then, contrastive learning is introduced into domain adaptation process. 
In particular, a pixel on the target domain is pulled closer to its corresponding prototype with the same class as its estimated pseudo-label and pushed away from other prototypes. In addition, in order to be invariant to domains, category-wise prototypes would be further updated by the current features of two domains. Besides, such prototypical contrastive adaptation scheme is applied at the feature and output level simultaneously. Based on the self-training framework, we further improve the performance with class-aware pseudo-label thresholds. 

Experimental results on the domain adaptation benchmarks for semantic segmentation, {\em i.e., GTA5 $\to$ Cityscapes and SYNTHIA $\to$ Cityscapes} further demonstrate the effectiveness of our approach, leading to the state-of-the-art performance. Specifically, with the DeepLab-v2 networks and ResNet-101 backbone, we achieve Cityscapes~\cite{cordts2016cityscapes} semantic segmentation mIoU by 56.3\% and 53.0\% when adapting from GTA5~\cite{richter2016playing} and SYNTHIA~\cite{ros2016synthia} datasets, largely outperforming previous state-of-the-arts.


We summarize the major contributions as follows:
\begin{itemize}
\item We propose {\em Prototypical Contrastive Adaptation} (\textit{ProCA}) by explicitly introducing constraints on features of different categories for UDA problem in semantic segmentation. 
This is implemented by not only pulling closer to prototypes with the same class, but also pushing away from prototypes with different classes simultaneously. A multi-level variant is also designed 
to further improve the adaptation ability.

\item Online prototypes updating scheme is introduced to improve the domain invariance and discriminant ability of class-wise prototypes. 

\item Combined with self-training method of class-wise adaptive thresholds, the proposed method achieves 56.3\% and 52.6\% mIoU when adapting GTA5 and SYNTHIA to Cityscapes, respectively, which outperforms previous state-of-the-arts by a large margin.
\end{itemize}
\section{Related Works}

\subsection{Semantic Segmentation}
Semantic segmentation is a fundamental computer vision task, which requires per-pixel predictions for a given image. Recently, with the help of convolution neural networks~\cite{long2015fully}, semantic segmentation has achieved remarkable progress. Numerous approaches focus to enlarge receptive fields~\cite{chen2017deeplab} and capture context information~\cite{zhao2017pyramid}. These methods generally require dense pixel-wise annotation datasets, such as Cityscapes~\cite{cordts2016cityscapes}, PASCAL VOC~\cite{everingham2010pascal} and ADE20K~\cite{zhou2017scene}. Since per-pixel level annotation of large amounts of data is time-consuming and expensive, some synthetic datasets are proposed such as GTA5~\cite{richter2016playing} and SYNTHIA~\cite{ros2016synthia} to generate largely labeled segmentation datasets at lower cost. However, when testing models trained on the synthetic datasets on the real-world datasets, significant performance drops are observed even for state-of-the-art methods. In presence of the domain shifts, we deal with the semantic segmentation task that aims to learn a well performing model on the target domain with only the source domain supervision.

\subsection{UDA for Semantic Segmentation}
Existing approaches for UDA of semantic segmentation can be primarily divided into three groups, including style transfer~\cite{murez2018image}, feature alignment~\cite{hoffman2018cycada,hoffman2016fcns,gu2021pit,zhou2021self} and self-training~\cite{zou2018unsupervised,araslanov2021self}. Motivated by the recent progress of unpaired image-to-image translation works~\cite{zhu2017unpaired}, researches on style transfer aim to learn the mapping from virtual to realistic data~\cite{murez2018image,hoffman2018cycada}. Previous works on feature alignment minimize the discrepancy between source and target domains to obtain domain-invariant features. This can be achieved by directly minimizing the Maximum Mean Discrepancy (MMD) distances across domains over domain-specific layers~\cite{long2015learning} or using discriminator to train the model in an adversarial way to avoid generating domain-aware discriminative features~\cite{hoffman2016fcns}. There are also some works attempting to absorb class-wise information into feature alignment. The fine-grained adversarial learning framework~\cite{wang2020classes} is proposed to incorporate class information into the discriminator, which helps to align feature in a class-aware manner, resulting better feature adaptation and performance. Approaches on self-training mainly focus on assigning pseudo-labels on target domain. Iterative self-training method is proposed~\cite{zou2018unsupervised} by alternatively generating pseudo-labels and retraining the model with a sampling module to deal with the category imbalanced issue. Uncertainty estimation~\cite{zheng2021rectifying} is proposed to rectify pseudo-label generation. Consistency based methods~\cite{araslanov2021self} have been adopted by enforcing consistency between predictions of different perturbations. In the work of~\cite{zhang2021prototypical}, a prototype-based sample-wise pseudo-label correction scheme is proposed and embeded into a complicated multi-stage training framework to enhance segmentation performance. Nevertheless, the methods above neglect the explicit modeling of the relationship between clusters of different categories, on the contrary, we directly explore such constraints of different category centroids by prototypical contrastive adaptation. In this way, the categories with similar distributions on the target domain can be easier to distinguish, leading to superior performance.

\subsection{Contrastive Learning}
Contrastive learning~\cite{he2020momentum, chen2020simple,zhou2021domain} has lead remarkable performance in self-supervised representation learning. STC~\cite{jiang2022stc} uses contrastive learning to learn association embeddings for video instance segmentation task. For UDA semantic segmentation, CLST~\cite{marsden2021contrastive} attempts to leverage contrastive learning to learn finer adapted feature representation. The concurrent work SDCA~\cite{li2021semantic} proposes using high-order semantic information to conduct contrast adaptation for UDA segmentation, which we found that it is not necessary. In this paper, with the aid of contrastive learning, we explicitly model the relationships of pixel-wise features between different categories and domains to obtain domain-invariant representation for unsupervised domain adaptive semantic segmentation.

\section{Methodology}
By minimizing the distributional distance between the source and target domains, previous approaches aim to obtain the domain-invariant representations for domain adaptation problem. However, the inter-class structural relationship is insufficiently explored. As shown in Figure~\ref{matching} (a), after alignment within the intra-class across two domains, it could be much more challenging to distinguish different categories since the decision boundaries identified on source domain could hardly be maintained on the target domain. Therefore, we propose a novel \emph{category-aware prototypical contrast adaptation} which introduces multiple prototypes to explicitly model the intra-class and inter-class relationships in a contrastive manner. 

Akin to previous state-of-the-art approaches~\cite{wang2020classes,zhang2021prototypical,li2021semantic}, a segmentation model is first trained on source domain in the supervised manner. Meanwhile, multiple prototypes are initialized to represent each category. Contrast adaptation is then adopted to constrain the inter-class relationship. Besides, prototypes are updated on both source domain and target domain to enhance the domain-invariant representations. As last, we present a modified pseudo-label generation method with class-aware adaptive thresholds for self-training, leading to new state-of-the-art performances. 



\subsection{Preliminaries}
Given the labeled source domain images $\mathcal{D}_{s} = \{(\mathbf{x}_n^s, y_n^s)\}_{n=1}^{N_s}$, as well as unlabeled target images $\mathcal{D}_{t} = \{(\mathbf{x}_n^{t})\}_{n=1}^{N_{t}}$, the goal of UDA of semantic segmentation is to train a model on $\mathcal{D}_s \text{ and } \mathcal{D}_{t}$; and evaluate the performance on the target domain. The segmentation model consists of a feature extractor $\mathcal{F}$ and a classifier $\mathcal{C}$, which predicts pixel-wise predictions for a given image.

Following previous works~\cite{hoffman2018cycada, wang2020classes, li2021semantic}, the segmentation model is first trained on the labeled source domain in a supervised manner by minimizing the loss between the prediction $p_n^s $ and the ground-truth label $Y_n^s \in \mathbb{L}^{H \times W}, \mathbb{L}=\{1,2,\cdots,C\}$ annotated with $C$ category labels, for a given image $x_n^s \in \mathbb{R}^{H \times W}$. We use the standard cross-entropy loss, which can be formulated as:

\begin{equation}
\label{ce_loss}
\mathcal{L}_n^{ce} = - \sum_{n=1}^{N_{s}} \sum_{i=1}^{H} \sum_{j=1}^{W} \sum_{c=1}^{C} y_{n,i,j,c}^s \log(p_{n,i,j,c}^s),
\end{equation}
where $N_s$ is the number of source domain images, $H$ and $W$ denote the height and the width of an image, $i$, $j$ are the pixel index of height and width, $C$ is the number of categories. $p_{n}^s \in \mathbb{R}^{H \times W \times C}$ is the predicted probability of the image $x_n^s$, which is obtained by up-sampling the prediction $\mathcal{C}(\mathcal{F}(x_n^s))$. $y_n^s \in \{0,1\}^{H \times W \times C}$ is the one-hot representation of the ground-truth label $Y_n^s$.

\subsection{Prototypical Contrast Adaptation}
\begin{figure*}[t]
\centering
\includegraphics[width=0.80\textwidth]{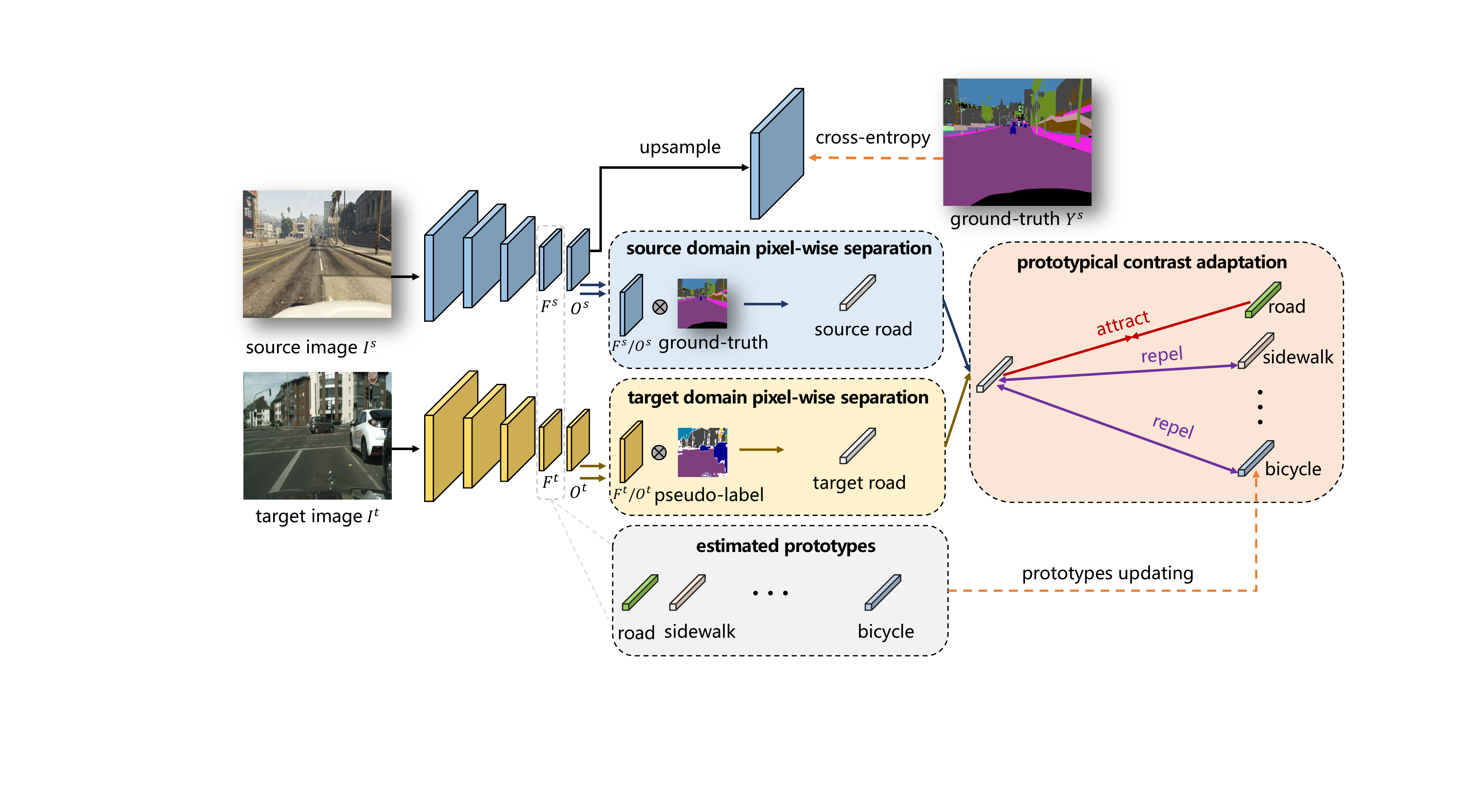}
\caption{The framework of proposed ProCA. For given source image $I^s$ and target image $I^t$, features $F^s$ and $F^t$ of two domains are first obtained through a shared feature encoder $\mathcal{F}$. Then, outputs $O^s$ and $O^t$ are obtained by a shared classifier $\mathcal{C}$. After obtaining initialized prototypes, a pixel from two domains acts as a contrastive manner with class-aware prototypes to directly model inter-class constraints. We conduct such prototypical contrast adaption on both feature-level and output-level. At last, the initialized prototypes are also updated during training to enhance the domain-invariant representational ability. }
\label{framework}
\end{figure*}

Here, intra-class and inter-class relations are simultaneously considered by prototypes-based contrastive learning as shown in Figure~\ref{framework}. Specifically, {\em ProCA} contains three stages, including {\em prototypes initialization}, {\em contrast adaptation} and {\em prototypes updating}.

\noindent \textbf{Prototypes Initialization}. After obtaining the model trained on the labeled source domain, the initialized class-aware prototypes can be calculated as:
\begin{equation}
\label{init_prototypes_feat_src}
\mathbf{p}_c^{feat} = \frac{\sum_{n=1}^{N_s} \sum_{i=1}^{H} \sum_{j=1}^{W} F_{n,i,j}^{s} \mathbbm{1}[Y_{n,i,j}^{s} = c]}{\sum_{n=1}^{N_s} \sum_{i=1}^{H} \sum_{j=1}^{W} \mathbbm{1}[Y_{n,i,j}^{s}=c]},
\end{equation}

where $F_{n,i,j}^s \in \mathbb{R}^{d}$ is the extracted source feature vector with dimension $d$, $c$ is the index of categories number $C$, $H$ and $W$ denote the height and width of the features, $\mathbbm{1}[Y_{n,i,j}^{s}=c]$ is an indicator function, which equals to $1$ if $Y_{n,i,j}^{s}=c$ and $0$ otherwise. Prototypes could be regarded as the approximated representational centroid of various categories.

\noindent \textbf{Contrast Adaptation}. Given an image of target domain, the corresponding feature $F_n^t$ is extracted by the shared backbone network $\mathcal{F}$. Accordingly, its pseudo-label $\widetilde y_n^t \in \{0,1\}^{H \times W \times C}$ could be produced by the classifier $\mathcal{C}$ trained on source domain. Here, pseudo-label could bridge the extracted features and their corresponding prototypes. Therefore, we could compute the similarity between features and each of prototypes, leading to a vector $P_{n,i,j}^{t\to s} = [P_{n,i,j, 1}^{t\to s}, \dots, P_{n,i,j,C}^{t\to s}]$:
\begin{equation}
\label{source-to-target}
 P_{n,i,j,c}^{t\to s} = \frac{\exp(\mathbf{p}_c^{feat} \cdot F_{n,i,j}^t / \tau)}
 {\sum_{c=1}^C\exp( \mathbf{p}_c^{feat} \cdot F_{n,i,j}^t / \tau )},
\end{equation}
where $\tau$ is the temperature. 
Then, we minimize the cross entropy loss between $P_{n,i,j}^{t\to s}$ and pseudo-label $\widetilde y_n^t$ as:
\begin{equation}
\label{s2t_loss}
\mathcal{L}_{n}^{t \to s} = - \sum_{i=1}^{H} \sum_{j=1}^{W} \sum_{c=1}^{C}  \widetilde y^t_{n,i,j,c} \log P_{n,i,j,c}^{s\to t}.
\end{equation}
The goal of such objective is to enforce the pixels belonging to the same category are supposed to share high representational similarity. 
In addition to the cross-domain adaptation, we also use source-source contrastive loss $\mathcal{L}_{n}^{s \to s}$ similarly:
\begin{equation}
\label{s2s_loss_feat}
\mathcal{L}_{n}^{s \to s} = - \sum_{i=1}^{H} \sum_{j=1}^{W} \sum_{c=1}^{C}  y^s_{n,i,j,c} \log P_{n,i,j,c}^{s\to s},
\end{equation}
where $y_n^s$ is the ground-truth one-hot source domain label, $P_{n,i,j}^{s \to s}$ is calculated similarly as Equation~\ref{source-to-target}. The final pixel-prototypes contrastive loss on feature-level is:

\begin{equation}
 \mathcal{L}_\mathrm{ContraFeat} = \sum_{n=1}^{N_t} \mathcal{L}_{n}^{t \to s} + \sum_{n=1}^{N_{s}} \mathcal{L}_{n}^{s \to s}. 
\end{equation}

\noindent \textbf{Prototypes Updating.} To enhance the domain-invariant representational ability of prototypes, we propose two schemes of prototype updating along with training to incorporate target-related information into prototypes. One is to update according to the computation of strict statistical mean of global data as:
\begin{equation}
\mathbf{p}_c^{feat} \leftarrow \frac{\mathbf{p}_c^{feat} n_c^{feat} + \mathbf{\widetilde p}_c^{feat} \widetilde n_c^{feat} }{n_c^{feat} + \widetilde n_c^{feat}},
\label{normal_updating}
\end{equation}
where $n_c^{feat}$ represents the accumulated number of pixels belonging to category $c$ until the last update,  $\mathbf{\widetilde p}_c^{feat}$ represents the online estimated prototypes for category $c$, and $\widetilde n_c^{feat}$ represents the total number of pixels belonging to category $c$ from a newly appended mini-batch during training.

In addition to source domain class-wise prototypes, we also leverage target features to update prototypes during feature adaptation process. This mixed prototypes scheme could be regarded as a bridge across two domains, which could naturally interact with each other. Thus, we further propose an alternative and more stable and robust way to directly update prototypes with a mixed domain scheme:
\begin{equation}
\mathbf{p}_c^{feat} \leftarrow m \mathbf{p}_c^{feat^{s}}  + (1 - m) \mathbf{p}_c^{feat^{t}},
\label{mixed_updating}
\end{equation}
where $m$ is a hyper-parameter, which defines a constant rate of source and target prototypes updating during training. $\mathbf{p}_c^{feat^{s}}$ is the estimated source prototype, and $\mathbf{p}_c^{feat^{t}}$ is the estimated target prototype.

\noindent \textbf{Label Space Adaptation.} 
As we mentioned before, prototypes are initialized, calculated and updated at the feature level \emph{i.e.}, output of the backbone network $\mathcal{F}$. Apart from this, we could also apply the proposed prototypical contrast adaptation in the label space \emph{i.e.}, the output of the classifier $\mathcal{C}$. The major difference is that the dimension of prototypes becomes the number of categories rather than the hidden channels in the feature space. Accordingly, the overall prototypical contrast adaptation losses becomes:
\begin{equation}
 \mathcal{L}_\mathrm{Contra} =  \mathcal{L}_\mathrm{ContraFeat} + \mathcal{L}_\mathrm{ContraOut}.
\end{equation}

\subsection{Combining ProCA with Self-Training}
\label{st}
Since the proposed category-aware prototypical contrast adaptation is orthogonal to self-training based methods, we further improve the adaptation performance through the self-training strategy following previous works~\cite{mei2020instance, li2021semantic}. 

\subsubsection{Class-wise Adaptive Pseudo-Label Thresholds.} After the prototypical contrast adaptation stage, we could obtain the sorted predicted confidence set $\theta_c = [\theta_{c,1},\theta_{c,2},...,\theta_{c,l_c}]$ of each category $c$, the length of confidence set belonging to category $c$ can be calculated as follows:
\begin{equation}
\label{num_pixels}
\mathbf{l}_c = \sum_{n=1}^{N_t} \sum_{i=1}^{H} \sum_{j=1}^{W} \mathbbm{1}[\widetilde Y_{n,i,j}^{t} = c],
\end{equation}
where $\widetilde Y_{n,i,j}^t \in \mathbb{L}^{H \times W}, \mathbb{L}=\{1,2,\cdots,C\}$ is the predicted pseudo-label for the image $x_n^t$. Then, each class threshold of pseudo-labels can be obtained by fixed percentage of the ranked confidence sets, where the percentage is denoted as a hyper-parameter $\eta$.

In addition to above self-training strategy, there are some works~\cite{zhang2021prototypical,zheng2021rectifying,mei2020instance} focusing on self-training itself improvements, like ProDA ~\cite{zhang2021prototypical} which leverages prototypes to obtain accurate pseudo-label. Since our proposed ProCA mainly works during feature adaptation process, which is orthogonal to such self-training based improvements. Thus we could combine our ProCA with such self-training methods to achieve better performance, which is shown in Table~\ref{different_st_method}.

\begin{table*}[thb]
\caption{Comparison results of \textbf{GTA5 $\to$ Cityscapes}. All methods use DeepLab-v2 with ResNet-101 backbone for \textbf{fair comparison}. $^\dagger$ means that we report the first stage self-training result of ProDA~\cite{zhang2021prototypical} for fair comparison, please see Table 3 of ProDA~\cite{zhang2021prototypical} for details.}
\centering
\resizebox{\textwidth}{!}{
\begin{tabular}{cc|ccccccccccccccccccc|cc}
\toprule
Method & 
Venue &
\rotatebox{90}{road} & 
\rotatebox{90}{sidewalk} & 
\rotatebox{90}{building} &
\rotatebox{90}{wall} &
\rotatebox{90}{fence} &
\rotatebox{90}{pole} &
\rotatebox{90}{light} &
\rotatebox{90}{sign} &
\rotatebox{90}{vegetation} &
\rotatebox{90}{terrain} &
\rotatebox{90}{sky} &
\rotatebox{90}{person} &
\rotatebox{90}{rider} &
\rotatebox{90}{car} &
\rotatebox{90}{trunk} &
\rotatebox{90}{bus} &
\rotatebox{90}{train} &
\rotatebox{90}{motorbike} &
\rotatebox{90}{bike} &
mIoU  & gain \\
\midrule
Source Only & - & 53.8 & 15.6 & 69.3 & 28.1 & 18.8 & 27.6 & 34.9 & 18.2 & 82.5 & 27.8 & 71.6 & 59.4 & 35.3 & 44.1 & 25.9 & 37.5 & 0.1 & 28.9 & 24.9 & 37.3 & +0.0\\
\midrule
PatchAlign~\cite{tsai2019domain} & CVPR'19 & 92.3 & 51.9 & 82.1 & 29.2 & 25.1 & 24.5 & 33.8 & 33.0 & 82.4 & 32.8 & 82.2 & 58.6 & 27.2 & 84.3 & 33.4 & 46.3 & 2.2 & 29.5 & 32.3 & 46.5  & +9.2 \\
ADVENT~\cite{vu2019advent} & CVPR'19 & 89.4 & 33.1 & 81.0 & 26.6 & 26.8 & 27.2 & 33.5 & 24.7 & 83.9 & 36.7 & 78.8 & 58.7 & 30.5 & 84.8 & 38.5 & 44.5 & 1.7 & 31.6 & 32.4 & 45.5  & +8.2 \\
BDL~\cite{li2019bidirectional} & CVPR'19 & 91.0 & 44.7 & 84.2 & 34.6 & 27.6 & 30.2 & 36.0 & 36.0 & 85.0 & 43.6 & 83.0 & 58.6 & 31.6 & 83.3 & 35.3 & 49.7 & 3.3 & 28.8 & 35.6 & 48.5  & +11.2 \\
UIDA~\cite{pan2020unsupervised} & CVPR'20 & 90.6 & 37.1 & 82.6 & 30.1 & 19.1 & 29.5 & 32.4 & 20.6 & 85.7 & 40.5 & 79.7 & 58.7 & 31.1 & 86.3 & 31.5 & 48.3 & 0.0 & 30.2 & 35.8 & 46.3  & +9.0 \\
LTIR~\cite{kim2020learning} & CVPR'20 & 92.9 & 55.0 & 85.3 & 34.2 & 31.1 & 34.9 & 40.7 & 34.0 & 85.2 & 40.1 & 87.1 & 61.0 & 31.1 & 82.5 & 32.3 & 42.9 & 0.3 & 36.4 & 46.1 & 50.2 & +12.9 \\
PIT~\cite{lv2020cross} & CVPR'20 & 87.5 & 43.4 & 78.8 & 31.2 & 30.2 & 36.3 & 39.9 & 42.0 & 79.2 & 37.1 & 79.3 & 65.4 & 37.5 & 83.2 & \textbf{46.0} & 45.6 & 25.7 & 23.5 & 49.9 & 50.6 & +13.2 \\
LSE~\cite{subhani2020learning} & ECCV'20 & 90.2 & 40.0 & 83.5 & 31.9 & 26.4 & 32.6 & 38.7 & 37.5 & 81.0 & 34.2 & 84.6 & 61.6 & 33.4 & 82.5 & 32.8 & 45.9 & 6.7 & 29.1 & 30.6 & 47.5 & +10.2 \\
WeakSeg~\cite{paul2020domain} & ECCV'20 & 91.6 & 47.4 & 84.0 & 30.4 & 28.3 & 31.4 & 37.4 & 35.4 & 83.9 & 38.3 & 83.9 & 61.2 & 28.2 & 83.7 & 28.8 & 41.3 & 8.8 & 24.7 & 46.4 & 48.2 & +10.9 \\
CrCDA~\cite{huang2020contextual} & ECCV'20 & 92.4 & 55.3 & 82.3 & 31.2 & 29.1 & 32.5 & 33.2 & 35.6 & 83.5 & 34.8 & 84.2 & 58.9 & 32.2 & 84.7 & 40.6 & 46.1 & 2.1 & 31.1 & 32.7 & 48.6 & +11.3 \\
FADA~\cite{wang2020classes} & ECCV'20 & 92.5 & 47.5 & 85.1 & 37.6 & \textbf{32.8} & 33.4 & 33.8 & 18.4 & 85.3 & 37.7 & 83.5 & 63.2 & 39.7 & 87.5 & 32.9 & 47.8 & 1.6 & 34.9 & 39.5 & 49.2 & +11.9 \\
IAST~\cite{mei2020instance} & ECCV'20 & \textbf{94.1} & \textbf{58.8} & 85.4 & 39.7 & 29.2 & 25.1 & 43.1 & 34.2 & 84.8 & 34.6 & 88.7 & 62.7 & 30.3 & 87.6 & 42.3 & \textbf{50.3} & 24.7 & 35.2 & 40.2 & 52.2 & +14.9 \\
ASA~\cite{zhou2020affinity} & TIP'21 & 89.2 & 27.8 & 81.3 & 25.3 & 22.7 & 28.7 & 36.5 & 19.6 & 83.8 & 31.4 & 77.1 & 59.2 & 29.8 & 84.3 & 33.2 & 45.6 & 16.9 & 34.5 & 30.8 & 45.1 & +7.8 \\
CLAN~\cite{luo2021category} & TPAMI'21 & 88.7 & 35.5 & 80.3 & 27.5 & 25.0 & 29.3 & 36.4 & 28.1 & 84.5 & 37.0 & 76.6 & 58.4 & 29.7 & 81.2 & 38.8 & 40.9 & 5.6 & 32.9 & 28.8 & 45.5 & +8.2 \\
DACS~\cite{tranheden2021dacs} & WACV'21 & 89.9 & 39.7 & \textbf{87.9} & 39.7 & 39.5 & 38.5 & 46.4 & 52.8 & \textbf{88.0} & \textbf{44.0} & \textbf{88.8} & 67.2 & 35.8 & 84.5 & 45.7 & 50.2 & 0.0 & 27.3 & 34.0 & 52.1 & +14.8 \\
RPLL~\cite{zheng2021rectifying} & IJCV'21 & 90.4 & 31.2 & 85.1 & 36.9 & 25.6 & 37.5 & \textbf{48.8} & 48.5 & 85.3 & 34.8 & 81.1 & 64.4 & 36.8 & 86.3 & 34.9 & 52.2 & 1.7 & 29.0 & 44.6 & 50.3 & +13.0 \\
DAST~\cite{yu2021dast} & AAAI'21 & 92.2 & 49.0 & 84.3 & 36.5 & 28.9 & 33.9 & 38.8 & 28.4 & 84.9 & 41.6 & 83.2 & 60.0 & 28.7 & 87.2 & 45.0 & 45.3 & 7.4 & 33.8 & 32.8 & 49.6 & +12.3 \\
ConTrans~\cite{lee2020unsupervised} & AAAI'21 & 95.3 & 65.1 & 84.6 & 33.2 & 23.7 & 32.8 & 32.7 & 36.9 & 86.0 & 41.0 & 85.6 & 56.1 & 25.9 & 86.3 & 34.5 & 39.1 & 11.5 & 28.3 & 43.0 & 49.6 & +13.2 \\
CIRN~\cite{gao2021addressing} & AAAI'21 & 91.5 & 48.7 & 85.2 & 33.1 & 26.0 & 32.3 & 33.8 & 34.6 & 85.1 & 43.6 & 86.9 & 62.2 & 28.5 & 84.6 & 37.9 & 47.6 & 0.0 & 35.0 & 36.0 & 49.1 & +11.8 \\
SDCA~\cite{li2021semantic} & Arxiv'21 & 92.8 & 52.5 & 85.9 & 34.8 & 28.1 & 40.3 & 44.4 & 33.4 & 86.7 & 41.7 & 87.1 & 67.4 & 37.3 & 88.1 & 39.9 & 52.5 & 1.4 & 34.2 & 55.0 & 52.9 & +15.6 \\
PWCL~\cite{liu2021domain} & Arxiv'21 & 93.3 & 54.2 & 83.0 & 25.9 & 28.1 & 37.2 & 41.1 & 39.3 & 83.1 & 38.9 & 78.2 & 61.3 & 36.2 & 84.2 & 35.8 & 54.0 & 18.1 & 26.7 & 47.5 & 50.9 & +13.6 \\
CLST~\cite{marsden2021contrastive} & Arxiv'21 & 92.8 & 53.5 & 86.1 & 39.1 & 28.1 & 28.9 & 43.6 & 39.4 & 84.6 & 35.7 & 88.1 & 63.9 & 38.3 & 86.0 & 41.6 & 50.6 & 0.1 & 30.4 & 51.7 & 51.6 & +14.3 \\
ESL~\cite{saporta2020esl} & CVPR'21 & 90.2 & 43.9 & 84.7 & 35.9 & 28.5 & 31.2 & 37.9 & 34.0 & 84.5 & 42.2 & 83.9 & 59.0 & 32.2 & 81.8 & 36.7 & 49.4 & 1.8 & 30.6 & 34.1 & 48.6 & +11.3 \\
MetaCorrect~\cite{guo2021metacorrection} & CVPR'21  & 92.8 & 58.1 & 86.2 & 39.7 & 33.1 & 36.3 & 42.0 & 38.6 & 85.5 & 37.8 & 87.6 & 62.8 & 31.7 & 84.8 & 35.7 & 50.3 & 2.0 & 36.8 & 48.0 & 52.1 & +14.8 \\
ProDA$^\dagger$~\cite{zhang2021prototypical} & CVPR'21 & 91.5 & 52.3 & 82.9 & \textbf{42.0} & 35.7 & 40.0 & 44.4 & \textbf{43.2} & 87.0 & 43.8 & 79.5 & 66.4 & 31.3 & 86.7 & 41.1 & 52.5 & 0.0 & \textbf{45.4} & 53.8  & 53.7 & +16.4 \\
UPLR~\cite{wang2021uncertainty} & ICCV'21 & 90.5 & 38.7 & 86.5 & 41.1 & 32.9 & 40.5 & 48.2 & 42.1 & 86.5 & 36.8 & 84.2 & 64.5 & 38.1 & 87.2 & 34.8 & 50.4 & 0.2 & 41.8 & 54.6 & 52.6 & +15.3 \\
\midrule
\emph{Ours} & - & 91.9 & 48.4 & 87.3 & 41.5 & 31.8 & \textbf{41.9} & 47.9 & 36.7 & 86.5 & 42.3 & 84.7 & \textbf{68.4} & \textbf{43.1} & \textbf{88.1} & 39.6 & 48.8 & \textbf{40.6} & 43.6 & \textbf{56.9} & \textbf{56.3} & \textbf{+19.0} \\
\bottomrule
\end{tabular}}
\label{gta2city}
\end{table*}

\section{Experiments}
\noindent \textbf{Datasets and Evaluation Metrics:}
Following previous works~\cite{wang2020classes,li2021semantic}, we evaluate the model in common UDA of semantic segmentation benchmarks, GTA5~\cite{richter2016playing} $\to$ Cityscapes~\cite{cordts2016cityscapes} and SYNTHIA~\cite{ros2016synthia} $\to$ Cityscapes~\cite{cordts2016cityscapes}. GTA5 is an image dataset synthesized by a photo-realistic open-world computer game. which shares 19 classes with Cityscapes. It has 24,966 images with the resolution $1914$ $\times$ $1052$. SYNTHIA is a synthetic urban scene dataset. Following previous works~\cite{tsai2018learning}, we use the subset SYNTHIA-RAND-CITYSCAPES sharing 16 common classes with Cityscapes. It contains 9400 images with the resolution $1280$ $\times$ $760$. Cityscapes is a dataset of real urban scenes, which is collected from 50 cities in Germany and neighboring cities. It has 2,975 training images, 500 validation images, and 1,525 test images, with the resolution 2048 $\times$ 1024. We report the results on Cityscapes validation set using the category-wise Intersection over Union (IoU). Specifically, we report the mean IoU (mIoU) of all 19 classes in GTA5 $\to$ Cityscapes setting and the 16 common categories in SYNTHIA $\to$ Cityscapes setting. In addition, since some works~\cite{tsai2018learning, luo2021category} only report mIoU for 13 common categories in SYNTHIA $\to$ CItyscapes setting, we also report the 13 common categories performance denoted as mIoU*.

\noindent \textbf{Implementation Details.} Following most previous works~\cite{hoffman2018cycada, wang2020classes, li2021semantic}, we use the DeepLab-v2 framework~\cite{chen2017deeplab} with ResNet-101~\cite{he2016deep} encoder as our segmentation model for fair comparison. All models are pre-trained on ImageNet~\cite{deng2009imagenet}. Atrous Spatial Pyramid Pooling (ASPP)~\cite{chen2017deeplab} is inserted after the last encoder layer with dilated rates \{6, 12, 18, 24\}. At last, an up-sampling layer is used to obtain the final per-pixel predictions with the same image size as input. We implement the proposed method with PyTorch~\cite{paszke2019pytorch} on NVIDIA Tesla V100. We apply SGD optimizer with the initial learning rate of $2.5 \times 10^{-4}$, momentum 0.9 and weight decay of $5.0 \times 10^{-4}$. We use polynomial learning rate scheduling with the power of 0.9. During prototypical contrast adaptation, the pseudo-label threshold of target domain is set to 0.9. For self-training stage. we assign pseudo-labels based on the predicted category probabilities with the adaptive thresholds. The percentage $\eta$ of the number of pixels for each category is 0.6 in default.

\begin{table}[t!]
\caption{Ablation studies of each component for GTA5 $\to$ Cityscapes. F refers to feature-level prototypical contrast adaptation; O refers to output-level prototypical contrast adaptation; Ada-ST refers to adaptive threshold self-training; MST refers to multi-scale testing. All methods use DeepLab-v2 with ResNet-101 backbone.}
    \centering
    \begin{tabular}{c c c c c c }
    \toprule
         Source Only & F & O  & Ada-ST & MST &mIoU  \\
         \hline
         \cmark & & & &  & 37.3 \\
          \cmark & \cmark & & &  & 47.9 \\
         \cmark & & \cmark & & & 48.4 \\
         \cmark & \cmark & \cmark & & & 48.8 \\\hline
          \cmark & &  & \cmark & &  43.9 \\
         \cmark & \cmark & \cmark & \cmark & & 55.1 \\
         \cmark & \cmark & \cmark & \cmark & \cmark & \textbf{56.3} \\
         \bottomrule
    \end{tabular}
\label{diff_comp}
\end{table}

\begin{table}[t!]
    \caption{Ablation studies of different domain alignment methods for GTA5 $\to$ Cityscapes. FADA~\cite{wang2020classes} refers fine-grained adversarial training for feature-level and output-level; SDCA~\cite{li2021semantic} refers semantic distribution-aware adaptation; Memory Bank refers to pixel-level bank for contrast adaptation. ProCA refers to prototypical contrast adaptation. All methods use DeepLab-v2 with ResNet-101 backbone.}
    \centering
    \begin{tabular}{cccccc}
    \toprule
         Source Only &  FADA  & SDCA & Memory Bank & ProCA & mIoU  \\
         \hline
         \cmark & &  &  &  & 37.3 \\
         \cmark & \cmark & & & &  46.9 \\
         \cmark & & \cmark & & & 47.2 \\
         \cmark & & & \cmark & & 47.6 \\
         \cmark &  &  & & \cmark & \textbf{48.8} \\
         \bottomrule
    \end{tabular}
    \label{diff_method}
\end{table}

\subsection{Comparisons with State-of-the-Art Methods}

In order to compare with previous state-of-the-art methods comprehensively, we include two typical methods: 1) Domain alignment methods which aim to align the distribution between source and target domains by distribution distances or adversarial training, including LITR~\cite{kim2020learning}, PIT~\cite{lv2020cross}, WeakSeg~\cite{paul2020domain}, CrCDA~\cite{huang2020contextual}, FADA~\cite{wang2020classes}, ASA~\cite{zhou2020affinity}, CLAN~\cite{luo2021category}, ConTrans~\cite{lee2020unsupervised}, SDCA~\cite{li2021semantic}, and CIRN~\cite{gao2021addressing}. 2) Self-training approaches, including UIDA~\cite{pan2020unsupervised}, LSE~\cite{subhani2020learning}, IAST~\cite{mei2020instance}, DACS~\cite{tranheden2021dacs}, RPLL~\cite{zheng2021rectifying}, DAST~\cite{yu2021dast}, ESL~\cite{saporta2020esl}, MetaCorrect~\cite{guo2021metacorrection}, and ProDA~\cite{zhang2021prototypical}.

\noindent \textbf{Results on GTA5 $\to$ Cityscapes.}  As shown in Table ~\ref{gta2city}, our approach achieves 56.3~\% mIoU, outperforming prior methods by a large margin. In particular, the most challenging classes stated in ~\cite{li2021semantic} including pole, person, rider, bike, and train, obtains the significant improvements, compared to previous work. It demonstrates our motivation that the inter-class modeling via prototypes indeed help the category recognition on the target domain, especially for the harder classes. 


\begin{table*}[t!]
\caption{Comparison results of \textbf{SYNTHIA $\to$ Cityscapes}. mIoU* denotes the mean IoU of 13 classes, which excludes the classes marked by the asterisk. All methods use DeepLab-v2 with ResNet-101 backbone for fair comparison. $^\dagger$ means that we report the first stage self-training result of ProDA~\cite{zhang2021prototypical} for fair comparison. The result of ProDA~\cite{zhang2021prototypical} is from their released code.}
\centering
\resizebox{\textwidth}{!}{
\begin{tabular}{cc|cccccccccccccccc|cc|cc}
\toprule
Method &
Venue &
\rotatebox{90}{road} & 
\rotatebox{90}{sidewalk} & 
\rotatebox{90}{building} &
\rotatebox{90}{wall*} &
\rotatebox{90}{fence*} &
\rotatebox{90}{pole*} &
\rotatebox{90}{light} &
\rotatebox{90}{sign} &
\rotatebox{90}{vegetation} &
\rotatebox{90}{sky} &
\rotatebox{90}{person} &
\rotatebox{90}{rider} &
\rotatebox{90}{car} &
\rotatebox{90}{bus} &
\rotatebox{90}{motorbike} &
\rotatebox{90}{bike} &
mIoU &
gain &
mIoU* &
gain* \\
\bottomrule
Source Only & - & 55.6 & 23.8 & 74.6 & 9.2 & 0.2 & 24.4 & 6.1 & 12.1 & 74.8 & 79.0 & 55.3 & 19.1 & 39.6 & 23.3 & 13.7 & 25.0 & 33.5 & 0.0 & 38.6 & 0.0 \\
\bottomrule
PatchAlign~\cite{tsai2019domain} & CVPR'19 & 82.4 & 38.0 & 78.6 & - & - & - & 9.9 & 10.5 & 78.2 & 80.5 & 53.5 & 19.6 & 67.0 & 29.5 & 21.6 & 31.3 & - & - & 46.5 & +7.9 \\
ADVENT~\cite{vu2019advent} & CVPR'19 & 85.6 & 42.2 & 79.7 & 8.7 & 0.4 & 25.9 & 5.4 & 8.1 & 80.4 & 84.1 & 57.9 & 23.8 & 73.3 & 36.4 & 14.2 & 33.0 & 41.2 & +7.7 & 48.0 & +9.4 \\
BDL~\cite{li2019bidirectional} & CVPR'19 & 86.0 & 46.7 & 80.3 & - & - & - & 14.1 & 11.6 & 79.2 & 81.3 & 54.1 & 27.9 & 73.7 & 42.2 & 25.7 & 45.3 & - & - & 51.4 & +12.8 \\
UIDA~\cite{pan2020unsupervised} & CVPR'20 & 84.3 & 37.7 & 79.5 & 5.3 & 0.4 & 24.9 & 9.2 & 8.4 & 80.0 & 84.1 & 57.2 & 23.0 & 78.0 & 38.1 & 20.3 & 36.5 & 41.7 & +8.2 & 48.9 & +10.3 \\
LTIR~\cite{kim2020learning} & CVPR'20 & 92.6 & 53.2 & 79.2 & - & - & - & 1.6 & 7.5 & 78.6 & 84.4 & 52.6 & 20.0 & 82.1 & 34.8 & 14.6 & 39.4 & - & - & 49.3 & +10.7 \\
PIT~\cite{lv2020cross} & CVPR'20 & 83.1 & 27.6 & 81.5 & 8.9 & 0.3 & 21.8 & 26.4 & 33.8 & 76.4 & 78.8 & 64.2 & 27.6 & 79.6 & 31.2 & 31.0 & 31.3 & 44.0 & +10.5 & 51.8 & +13.2 \\
LSE~\cite{subhani2020learning} & ECCV'20 & 82.9 & 43.1 & 78.1 & 9.3 & 0.6 & 28.2 & 9.1 & 14.4 & 77.0 & 83.5 & 58.1 & 25.9 & 71.9 & 38.0 & 29.4 & 31.2 & 42.6 & +9.1 & 49.4 & +10.8 \\
CrCDA~\cite{huang2020contextual} & ECCV'20 & 86.2 & 44.9 & 79.5 & 8.3 & 0.7 & 27.8 & 9.4 & 11.8 & 78.6 & 86.5 & 57.2 & 26.1 & 76.8 & 39.9 & 21.5 & 32.1 & 42.9 & +9.4 & 50.0 & +11.4 \\
WeakSeg~\cite{paul2020domain} & ECCV'20 & 92.0 & 53.5 & 80.9 & 11.4 & 0.4 & 21.8 & 3.8 & 6.0 & 81.6 & 84.4 & 60.8 & 24.4 & 80.5 & 39.0 & 26.0 & 41.7 & 44.3 & +10.8 & 51.9 & +13.3 \\
IAST~\cite{mei2020instance} & ECCV'20 & 81.9 & 41.5 & 83.3 & 17.7 & \textbf{4.6} & 32.3 & 30.9 & 28.8 & 83.4 & 85.0 & 65.5 & 30.8 & 86.5 & 38.2 & \textbf{33.1} & 52.7 & 49.8 & +16.3 & 57.0 & +18.4 \\
FADA~\cite{wang2020classes} & ECCV'20 & 84.5 & 40.1 & 83.1 & 4.8 & 0.0 & 34.3 & 20.1 & 27.2 & 84.8 & 84.0 & 53.5 & 22.6 & 85.4 & 43.7 & 26.8 & 27.8 & 45.2 & +11.7 & 52.5 & +13.9 \\
ASA~\cite{zhou2020affinity} & TIP'21 & 91.2 & 48.5 & 80.4 & 3.7 & 0.3 & 21.7 & 5.5 & 5.2 & 79.5 & 83.6 & 56.4 & 21.9 & 80.3 & 36.2 & 20.0 & 32.9 & 41.7 & +8.2 & 49.3 & +10.7 \\
CLAN~\cite{luo2021category} & TPAMI'21 & 82.7 & 37.2 & 81.5 & - & - & - & 17.1 & 13.1 & 81.2 & 83.3 & 55.5 & 22.1 & 76.6 & 30.1 & 23.5 & 30.7 & - & - & 48.8 & +10.2 \\
DACS~\cite{tranheden2021dacs} & WACV'21 & 80.6 & 25.1 & 81.9 & 21.5 & 2.9 & 37.2 & 22.7 & 24.0 & 83.7 & \textbf{90.8} & 67.6 & \textbf{38.3} & 82.9 & 38.9 & 28.5 & 47.6 & 48.3 & +14.8 & 54.8 & +16.2 \\
RPLL~\cite{zheng2021rectifying} & IJCV'21 & 87.6 & 41.9 & 83.1 & 14.7 & 1.7 & 36.2 & 31.3 & 19.9 & 81.6 & 80.6 & 63.0 & 21.8 & 86.2 & 40.7 & 23.6 & 53.1 & 47.9 & +14.4 & 54.9 & +16.3 \\
CIRN~\cite{gao2021addressing} & AAAI'21 & 85.8 & 40.4 & 80.4 & 4.7 & 1.8 & 30.8 & 16.4 & 18.6 & 80.7 & 80.4 & 55.2 & 26.3 & 83.9 & 43.8 & 18.6 & 34.3 & 43.9 & +10.4 & 51.1 & +12.5 \\
DAST~\cite{yu2021dast} & AAAI'21 & 87.1 & 44.5 & 82.3 & 10.7 & 0.8 & 29.9 & 13.9 & 13.1 & 81.6 & 86.0 & 60.3 & 25.1 & 83.1 & 40.1 & 24.4 & 40.5 & 45.2 & +11.7 & 52.5 & +13.9 \\
ConTrans~\cite{lee2020unsupervised} & AAAI'21 & \textbf{93.3} & \textbf{54.0} & 81.3 & 14.3 & 0.7 & 28.8 & 21.3 & 22.8 & 82.6 & 83.3 & 57.7 & 22.8 & 83.4 & 30.7 & 20.2 & 47.2 & 46.5 & +13.0 & 53.9 & +15.3 \\
SDCA~\cite{li2021semantic} & Arxiv'21 & 88.4 & 45.9 & 83.9 & 24.0 & 1.7 & 38.1 & 25.2 & 17.0 & 85.3 & 82.9 & 67.3 & 26.6 & 87.1 & 47.2 & 28.6 & 53.4 & 50.2 & +16.7 & 56.8 & +18.2 \\
PWCL~\cite{liu2021domain} & Arxiv'21 & - & - & - & - & - & - & - & - & - & - & - & - & - & - & - & - & - & - & 53.3 & +14.7 \\
CLST~\cite{marsden2021contrastive} & Arxiv'21 & 88.0 & 49.2 & 82.2 & 16.3 & 0.4 & 29.2 & 31.8 & 23.9 & 84.1 & 88.0 & 59.1 & 27.2 & 85.5 & 46.4 & 28.9 & \textbf{56.5} & 49.8 & +16.3 & 57.8 & +19.2 \\
ESL~\cite{saporta2020esl} & CVPR'21 & 84.3 & 39.7 & 79.0 & 9.4 & 0.7 & 27.7 & 16.0 & 14.3 & 78.3 & 83.8 & 59.1 & 26.6 & 72.7 & 35.8 & 23.6 & 45.8 & 43.5 & +10.0 & 50.7 & +12.1 \\
MetaCorrect~\cite{guo2021metacorrection} & CVPR'21 & 92.6 & 52.7 & 81.3 & 8.9 & 2.4 & 28.1 & 13.0 & 7.3 & 83.5 & 85.0 & 60.1 & 19.7 & 84.8 & 37.2 & 21.5 & 43.9 & 45.1 & +11.6 & 52.5 & +13.9 \\
ProDA$^\dagger$~\cite{zhang2021prototypical} & CVPR'21 & 87.1 & 44.0 & 83.2 & 26.9 & 0.0 & \textbf{42.0} & \textbf{45.8} & \textbf{34.2} & 86.7 & 81.3 & 68.4 & 22.1 & 87.7 & 50.0 & 31.4 & 38.6 & 51.9 & +18.4 & 58.5 & +19.9 \\
UPLR~\cite{wang2021uncertainty} & ICCV'21 & 79.4 & 34.6 & 83.5 & 19.3 & 2.8 & 35.3 & 32.1 & 26.9 & 78.8 & 79.6 & 66.6 & 30.3 & 86.1 & 36.6 & 19.5 & 56.9 & 48.0 & +14.5 & 54.6 & +16.0\\
\bottomrule
\emph{Ours} & - & 90.5 & 52.1 & \textbf{84.6} & \textbf{29.2} & 3.3 & 40.3 & 37.4 & 27.3 & \textbf{86.4} & 85.9 & \textbf{69.8} & 28.7 & \textbf{88.7} & \textbf{53.7} & 14.8 & 54.8 & \textbf{53.0} & \textbf{+19.5} & \textbf{59.6} & \textbf{+21.0} \\
\toprule
\end{tabular}}
\label{syn2city}
\end{table*}

\noindent \textbf{Results on SYNTHIA $\to$ Cityscapes.}
The comparisons of SYNTHIA $\to$ Cityscapes are shown in Table ~\ref{syn2city}. Among all the 16 categories, we achieve the best scores on 6 categories, most of those are hard classes stated in ~\cite{li2021semantic}, {\em e.g.,} person, and bike. To be specific, the proposed method achieves the mIoU score by 53.0\% and 59.6\% over the 16 and 13 categories respectively, which obtains the gains over the baseline by 19.5\% and 21.0\%.

\begin{table}[t!]
\caption{Ablation studies of different prototypes updating scheme for GTA5 $\to$ Cityscapes. Fixed refers to no-updating for calculated prototypes; Source means updating in a strict statistical way on in source domain as Equation~\ref{normal_updating}; Mixed refers updating in Equation~\ref{mixed_updating} in both source and target domain. 
}
    \centering
    \begin{tabular}{c c c c c }
    \toprule
         Source Only &  Fixed  & Source & Mixed & mIoU  \\
         \hline
         \cmark & &  &  & 37.3 \\
         \cmark & \cmark & & & 47.8 \\
         \cmark &  & \cmark & & 48.3 \\
         \cmark &  &  & \cmark & \textbf{48.8} \\
         \bottomrule
    \end{tabular}
\label{updating_method}
\end{table}

\begin{table}[t!]
\caption{Ablation studies of different self-training schemes for GTA5 $\to$ Cityscapes. Naive Self-Training refers to fixed 0.9 threshold for pseudo-label generation; Adaptive Self-Training refers to adaptive pseudo-label generation, which is median of predicted confidence set of each class in default (Sec \ref{st}). Prototypes-based Self-Training refers to pseudo-label generation strategy by utilizing prototypes which is proposed by ProDA~\cite{zhang2021prototypical}.
}
    \centering
    \begin{tabular}{ccccc}
    \toprule
         ProCA & Naive & Adaptive & Prototypes-based & mIoU  \\
         \hline
         \cmark & &  &  & 48.8 \\
         \cmark & \cmark & & & 55.2 \\
         \cmark &  & \cmark & & 56.3 \\
         \cmark &  &  & \cmark & 57.5 \\
         \bottomrule
    \end{tabular}
\label{different_st_method}
\end{table}

\noindent \textbf{Discussion with ProDA.}
It should be noticeable that our proposed prototype contrastive learning method surpass a similar prototype-based method ProDA~\cite{zhang2021prototypical} on both transferring scenarios under a \emph{fair} comparison setting. Especially, in GTA5 $\to$ Cityscapes, our adaptive method outperforms~\cite{zhang2021prototypical} by a large margin of $1.4\%$ mIOU. This is due to the fact that ProDA only utilizes prototypes to rectify pseudo-labels or align feature in a purely sample-wise manner, which is more vulnerable to the interference from outlier or noisy samples in the target domain, while our pipeline directly depicts the class-wise relation in a sample-to-prototype manner, making the learning process more robust and friendly to cross-domain transferring.  

\noindent \textbf{Discussion with Other Contrastive Learning based Methods.}
It should also be noticed that compared with a similar patch-wise contrastive learning method PWCL~\cite{liu2021domain}, our approach achieves superiority of $4.2\%$ and $5.4\%$ mIOU improvement on both GTA5 $\to$ Cityscapes and SYNTHIA $\to$ Cityscapes respectively.
This is due to the fact that PWCL only takes patch-wise features for contrastive feature adaptation, which is coarse to depict class-wise relation and ignores the fine-grained pixel-wise distribution variation during training process, resulting in less discriminative and general representation.

\noindent \textbf{How ProCA helps poor classes adaptation?} As shown in Table\ref{gta2city}, the performance of \emph{train} class could not be improved by state-of-the-art pseudo-label method ProDA. This is because initialized predictions are totally wrong, thus ProDA could not estimate accurate pseudo-label for \emph{train} class. Different from ProDA, our ProCA first corrects the \emph{train} class predictions by push aware from others class centroids, which progressively obtain more and more accurate feature representation of \emph{train} class. After introducing such relationship between different classes, our proposed method achieves highest \emph{train} class performance after combining with self-training method.

\subsection{Ablation Studies}

\noindent \textbf{Effectiveness of Each Component.} We conduct ablation studies to demonstrate the effectiveness of each component. We use the ResNet-101 backbone with DeepLab-v2 segmentation for GTA5 $\to$ Cityscapes adaptation. As shown in Table~\ref{diff_comp}, the source-only baseline achieves 37.3\% mIoU on Cityscapes val set. Further, we achieve 48.8\% mIoU score after using the proposed prototypical contrast adaptation. At last, the performance can be improved to 55.1\% mIoU through self-training with class-aware adaptive thresholds. Finally, we obtain 56.3\% mIoU score by multi-scale testing following FADA~\cite{wang2020classes}. When directly using self-training after source-domain training, we could only obtain 43.9\% mIoU, which is 11.2\% mIoU lower than 55.1\% mIoU score, demonstrating the effectiveness of ProCA. 

\noindent \textbf{Effectiveness of ProCA.} To verify the effectiveness of ProCA, we implement other feature alignment methods, {\em e.g}, class-wise adversarial training without inter-class modeling FADA~\cite{wang2020classes}, semantic-distribution modeling with category-wise information. As shown in Table~\ref{diff_method}, FADA improves the baseline to 46.9\% mIoU, which indicates the effectiveness of the adversarial training. SDCA~\cite{li2021semantic} obtains 47.2\% mIoU by considering semantic-aware feature alignment. Memory Bank obtains 47.6\% mIoU by introducing pixel-wise contrastive adaptation, which already achieves better performance than FADA and SDCA. Compared with above methods, our ProCA achieves the best mIoU score 48.8\%, which demonstrates the superiority of the proposed class-aware prototypical contrast adaptation than pixel-wise memory bank scheme.

\begin{table}[t]
\caption{Ablation studies of different percentages determining class-wise thresholds during self-training process for GTA5 $\to$ Cityscapes. All methods use DeepLab-v2 with ResNet-101 backbone.}
    \centering
    \begin{tabular}{c|ccccccc}
    \toprule
          $\eta$ (\%) & 30 & 40 & 50 & 60 & 70 & 80 & 90  \\
         \hline
         mIoU & 54.3 & 54.7 & 54.9 & 55.1 & 54.4 & 54.1 & 52.5 \\
         \bottomrule
    \end{tabular}
\label{diff_percentage}
\end{table}

\begin{table}[t]
\caption{Ablation studies of different contrastive adaptation choices for GTA5 $\to$ Cityscapes. $s \to s$ means Eq.~\ref{s2s_loss_feat} and $t \to s$ means Eq.~\ref{s2t_loss}}.
    \centering
    \begin{tabular}{ccccc}
    \toprule
         Source Only & $s\to s$ & $t \to s$ & mIoU  \\
         \hline
         \cmark & &  &  37.3 \\
         \cmark & \cmark & & 44.9 \\
         \cmark &  & \cmark & 46.8 \\
         \cmark & \cmark & \cmark & 48.8 \\
         \bottomrule
    \end{tabular}
\label{different_choice}
\end{table}

\noindent \textbf{Effectiveness of Mixed Updating.} We conduce ablation studies to verify the effectiveness of mixed updating for prototypes. As shown in \ref{updating_method}, a naive fixed prototype scheme only achieves 47.8\% mIoU, while centroid updating way only in source domain obtains 48.3\% mIoU, which has 0.5\% gain compared with fixed-prototype scheme. Mixed updating scheme achieves best 48.8\% mIoU score, which demonstrates the effectiveness of latest features during training. 

\noindent \textbf{Effectiveness of Multi-Level Adaptation.} We conduct ablation studies to verify the effectiveness of multi-level adaptation. The results are shown in Table~\ref{diff_comp}. when only using feature-level adaptation or output-level adaptation, we achieve 47.9\% mIoU and 48.4\% mIoU, respectively. After combining them, we obtain the best mIoU score 48.8\%, demonstrating the superiority of multi-level adaptation. 

\noindent \textbf{Effectiveness of In-Domain Contrastive Adaptation and Cross-Domain Contrastive Adaptation.} We conduct experiments to study the influence of different domain choices of prototypical contrastive adaptation. The results are shown in Table ~\ref{different_choice}. When only using source-to-source ProCA scheme, we could obtain 7.6\% mIoU improvement. When only using cross-domain ProCA scheme, we could obtain 9.5\% mIoU improvement. After combining both in-domain and cross-domain strategies, we finally obtain 48.8\% mIoU, which verifies the effectiveness of the proposed method.

\noindent \textbf{Effectiveness of Different Percentages for Adaptive Self-training.} We conduct experiments to study the influence of different percentages of pseudo-labels generation during self-training stage. The results are shown in Table ~\ref{diff_percentage}. Using 60 percentage to generate pseudo-labels, ProCA achieves the best mIoU 55.1\%. And larger percentages harm the performance.

\begin{figure*}[t]
\centering
\includegraphics[width=0.8\textwidth]{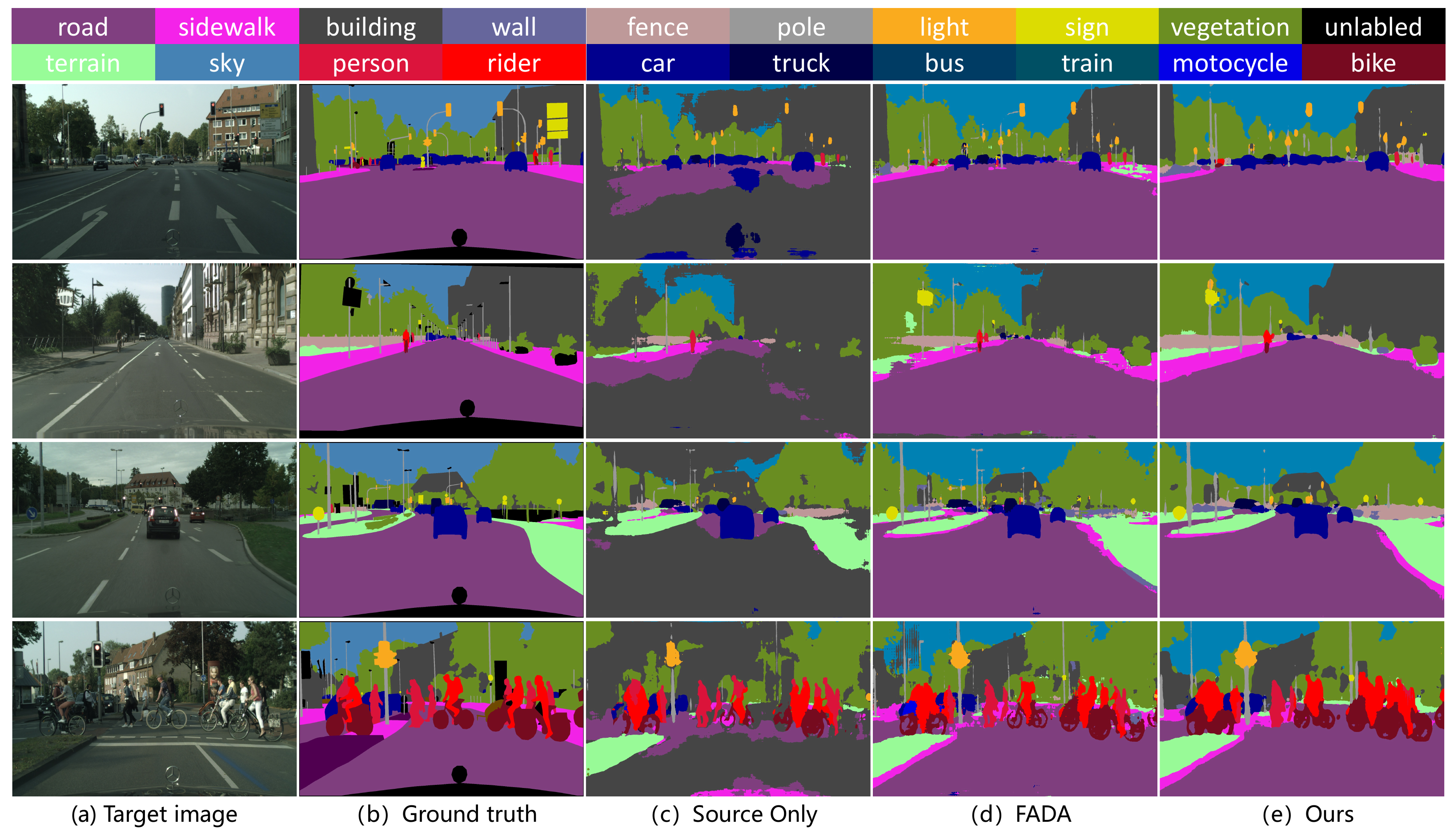}
\caption{Qualitative segmentation results for GTA5 $\to$ Cityscapes. From the left to right: target image, ground-truth, predictions by Source Only, FADA~\cite{wang2020classes} and our proposed method are shown.}
\label{visualize}
\end{figure*}

\section{Conclusions}
In this paper, we propose ProCA, which utilizes class-wise prototypes to align features in a fine-grained manner. Apart from feature-level adaptation, output-level prototypes are also exploited to boost the adaptation performance. The proposed method achieves the state-of-the-art performance on challenging benchmarks, outperforming previous methods by a large margin. Elaborate ablative studies demonstrate the advancement of our ProCA. We hope the proposed prototypical contrast adaptation could extend to more tasks, such as object detection and instance segmentation.

\clearpage
%
%
\bibliographystyle{splncs04}
\bibliography{egbib}
\end{document}